\documentclass[runningheads]{llncs}
\usepackage[T1]{fontenc}
\usepackage{graphicx,verbatim}
\usepackage{hyperref}
\usepackage{color}

\usepackage{algorithm}
\usepackage{algpseudocode}
\usepackage{amsmath}
\usepackage{amssymb}
\usepackage{booktabs}
\usepackage{tcolorbox}
\usepackage[pro]{fontawesome5}
\usepackage{xcolor}
\usepackage{colortbl}  
\usepackage{array}
\usepackage{multirow}

\definecolor{lightpurple}{HTML}{ebdef0}
\definecolor{lightblue}{HTML}{d6eaf8}
\definecolor{darkerblue}{HTML}{A9C4E2} 
\definecolor{lightgreen}{HTML}{d4efdf}
\definecolor{lightorange}{HTML}{fae5d3}
\definecolor{darkerorange}{HTML}{F0B27A} 
\definecolor{lightyellow}{HTML}{fcf3cf}
\definecolor{promptbg}{HTML}{ffe5ec} 
\definecolor{promptborder}{HTML}{ffb3c6} 
\definecolor{prompttitle}{HTML}{590d22} 
\definecolor{gold}{RGB}{255, 215, 0}

\tcbset{
    promptstyle/.style={
        colback=promptbg,
        colframe=promptborder,
        coltitle=prompttitle,
        fonttitle=\bfseries,
        boxrule=1mm,
        arc=4mm,
        left=3mm,
        right=3mm,
        top=3mm,
        bottom=3mm,
        title={},
        before skip=10pt,
        after skip=10pt,
        boxsep=5pt
    }
}

\begin{document}

\title{MOTOR: Multimodal Optimal Transport via Grounded Retrieval in Medical Visual Question Answering}
\titlerunning{MOTOR}

\author{Mai A. Shaaban\inst{1,2} \and
Tausifa Jan Saleem\inst{1} \and Vijay Ram Papineni\inst{3} \and
Mohammad Yaqub\inst{1}}

\authorrunning{Mai A. Shaaban et al.}
%
\institute{Mohamed bin Zayed University of Artificial Intelligence, Abu Dhabi, UAE
\\ \email{\{mai.kassem, tausifa.saleem, mohammad.yaqub\}@mbzuai.ac.ae} \and
Department of Mathematics and Computer Science, Faculty of Science, Alexandria University, Alexandria, EG
\\ \and
Sheikh Shakhbout Medical City, Abu Dhabi, UAE
\\}

%

\maketitle              
\begin{abstract}
Medical visual question answering (MedVQA) plays a vital role in clinical decision-making by providing contextually rich answers to image-based queries. Although vision-language models (VLMs) are widely used for this task, they often generate factually incorrect answers. Retrieval-augmented generation addresses this challenge by providing information from external sources, but risks retrieving irrelevant context, which can degrade the reasoning capabilities of VLMs. Re-ranking retrievals, as introduced in existing approaches, enhances retrieval relevance by focusing on query-text alignment. However, these approaches neglect the visual or multimodal context, which is particularly crucial for medical diagnosis. We propose MOTOR, a novel multimodal retrieval and re-ranking approach that leverages grounded captions and optimal transport. It captures the underlying relationships between the query and the retrieved context based on textual and visual information. Consequently, our approach identifies more clinically relevant contexts to augment the VLM input. Empirical analysis and human expert evaluation demonstrate that MOTOR achieves higher accuracy on MedVQA datasets, outperforming state-of-the-art methods by an average of 6.45\%. Code is available at \url{https://github.com/BioMedIA-MBZUAI/MOTOR}.


\end{abstract}

\section{Introduction}

\begin{figure}[t]
    \centering
    \includegraphics[width=\textwidth]{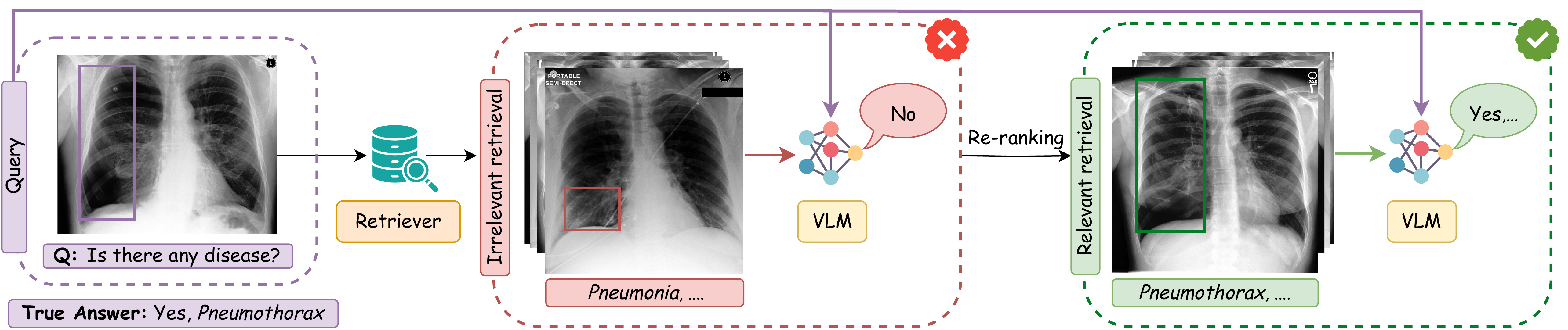}
    \caption{\textcolor{red}{Red}: misleading context is retrieved, causing an incorrect answer. \textcolor{green!70!black}{Green}: relevant context is prioritized by re-ranking initial retrievals, causing a correct answer.}
    \label{fig:example}
\end{figure}

Medical Visual Question Answering (MedVQA) is an intriguing AI task that entails answering natural language questions related to the medical domain based on images like MRIs, CT scans, X-rays, etc. These answers play a crucial role in the comprehensive analysis of medical images, enabling healthcare professionals to make accurate diagnoses. \cite{lin2023medical}. MedVQA has massive potential to support clinical decision-making, improve clinical efficiency, and enhance patient outcomes \cite{yang2023impact,thirunavukarasu2023large}. However, the requirement for high precision, domain-specific knowledge, and multimodal reasoning in medical tasks makes MedVQA highly complex \cite{zhang2023pmc}. \par
The ability of medical vision-language models (VLMs) to generate contextually relevant text based on medical imaging input has made them handy tools for MedVQA. These models produce remarkable outcomes across a range of benchmarks by utilizing large-scale multimodal pre-training \cite{chen2022multi,wang2022medclip,li2024llava,chen2024dragonfly}. However, standalone VLMs that operate independently without access to external knowledge sources or real-time data retrieval, encounter substantial difficulties, especially in low-resource settings. Due to inadequate context, they usually struggle with factual correctness and can generate hallucinatory or incomplete outputs \cite{zhang2023pmc,lewis2020retrieval,shi2024survey}.\par

Retrieval-augmented generation (RAG), which incorporates a retrieval component into the generative model, has demonstrated the potential to remedy the aforementioned disadvantages of standalone generation (SG) without additional training \cite{lewis2020retrieval,shi2024survey}. RAG includes pertinent context in the input by dynamically retrieving relevant information from external sources. This method has successfully filled in knowledge gaps in large language models (LLMs) and ensures that answers are based on contextually accurate information \cite{ragsurvey,gao2023retrieval}.\par

Most of the existing research in MedVQA focuses on improving SG by making architectural changes or applying different training strategies \cite{ragreview,ramm,xia2024rule}. Hence, the application of multimodal RAG (MM-RAG) remains in its infancy. Moreover, directly applying MM-RAG to VLMs poses the following challenges: (1) a limited number of retrieved elements might not encompass the reference knowledge needed to answer the question, thus restricting the faithfulness of the model \cite{xia2024rule}; (2) a large number of retrieved elements may include irrelevant and erroneous references, which can interfere with the model’s predictions \cite{shaaban2024medpromptx}. Research efforts have been directed towards leveraging rerankers to refine retrieved contexts \cite{rank_bm25,chen2024bge,li2023making}. The refinement of the ranking of elements retrieved during the initial retrieval process ensures that relevant contexts are prioritized, as depicted in Fig.~\ref{fig:example}. Nevertheless, existing rerankers are unimodal and focus solely on text-to-text alignment between the query and the retrieved elements, neglecting visual information. Despite the significance of multimodal alignment, it is inherently complex and requires advanced modelling techniques to capture the fine-grained underlying relationships \cite{zhang2023pmc}.\par

In this paper, we propose \textbf{M}ultimodal \textbf{O}ptimal \textbf{T}ransport via Gr\textbf{O}unded \textbf{R}etrieval (\textbf{MOTOR}), a novel retrieval and re-ranking approach specifically designed for the MM-RAG framework, tailored to MedVQA tasks. MOTOR first retrieves external images relevant to the query image from a database along with their corresponding medical reports. Then, the clinical relevance of retrievals is enhanced by integrating grounded captions (i.e., text descriptions with corresponding visual evidence) and optimal transport (OT) \cite{villani2009optimal,cuturi2013sinkhorn}, which supplements the reasoning capabilities of the VLM. To the best of our knowledge, we are the first to introduce multimodal re-ranking within medical MM-RAG frameworks for precise alignment, combining textual similarity and visual context. This integration enables robust and contextually accurate retrieval, significantly improving the factual accuracy of the generated answers. The main contributions are as follows:

\begin{itemize}
    \item We propose MOTOR, a novel training-free approach for retrieving precise contexts in a medical MM-RAG framework.
    \item We introduce a fine-grained visual-text alignment, which captures the underlying structures between the query and the retrieved elements, thereby improving clinical relevance.
    \item We perform automated and human expert evaluations across VLMs and MedVQA datasets to demonstrate the strength of our proposed approach.
\end{itemize}

\section{Methodology}

\begin{figure}[t]
\centering
\includegraphics[width=\textwidth]{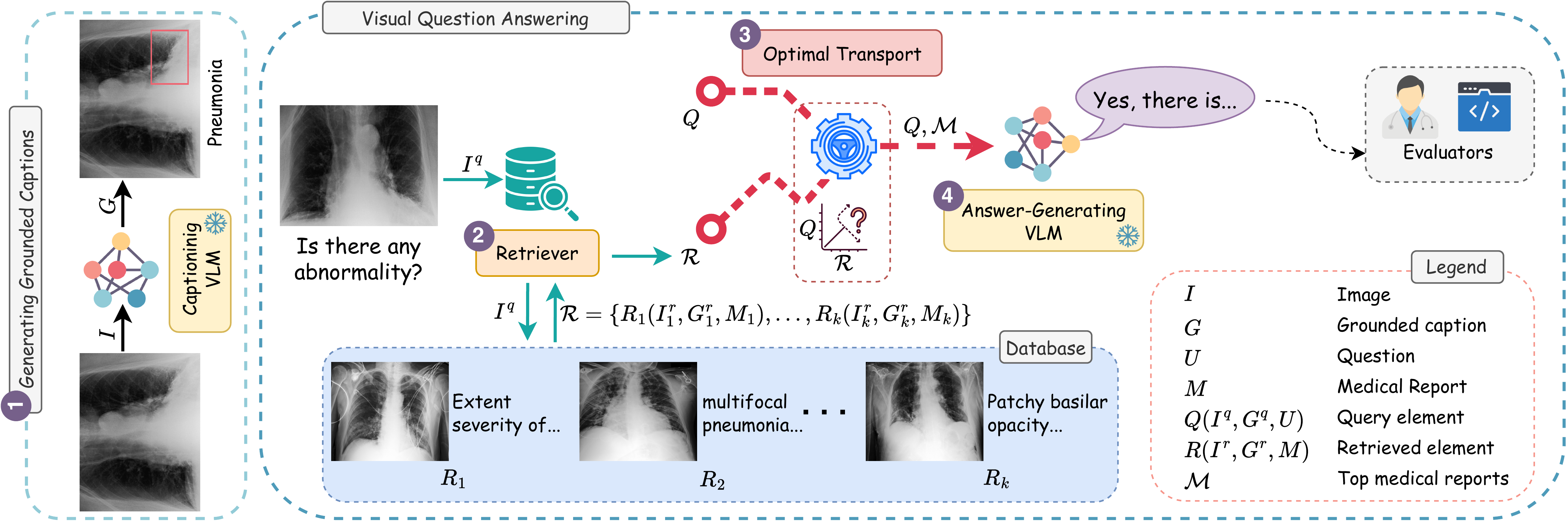}
\caption{MOTOR: 
(1) end-to-end generation of grounded captions for query images and database images; 
(2) retrieving top-$k$ similar elements using pre-computed image embeddings; 
(3) finding retrieved elements $R \in \mathcal{R}$ that have minimal OT cost w.r.t. $Q$; and (4) passing $Q$ and the most relevant reports $\mathcal{M}$ to a VLM to generate an answer.
}
\label{fig:method}
\end{figure}

This section provides a comprehensive outline of MOTOR, which is structured into the following phases: grounded caption generation, retrieval, OT-based re-ranking, and answer generation (refer to Fig.~\ref{fig:method} for illustration).

\noindent{\textbf{Grounded Caption Generation.}}
We introduce an end-to-end grounded caption generation task that focuses on generating descriptions of abnormalities and also localizing them within medical images using spatial annotations. We process an input image $I$ to generate a grounded caption \( G = \{(t_1, b_1), \dots, (t_n, b_n)\} \) by MAIRA-2 \cite{bannur2024maira}, where $t_i$ denotes the $i$-th abnormality description, and $b_i$ denotes the corresponding bounding box. $G$ is later utilized in the re-ranking step to compare the query image against candidate images to ensure clinical relevance.

\noindent{\textbf{Retrieval.}}
The retriever consists of a vector database $\mathcal{D}$ that stores image-text pairs. The retriever indexes the pre-computed embeddings of images and then retrieves top-$k$ similar elements $\mathcal{R}=\{R_1(I_1^r, G_1^r, M_1), \dots, R_k(I_k^r, G_k^r, M_k)\}$ using consine similarity w.r.t. the query $Q(I^q, G^q, U)$ , where $M_i$ is a medical report for the $i$-th retrieved element, $U$ is a question and $\mathcal{R} \subset \mathcal{D}$.

\noindent{\textbf{OT-based Re-ranking.}}
We propose multimodal OT for the re-ranking step to prioritize the most relevant elements before passing their medical reports to a VLM. Hence,  it addresses the reasoning gap in standard RAG, which stems from limited evidence scope and ungrounded context. OT is a paradigm that transforms one probability distribution into another at minimal cost \cite{villani2009optimal,cuturi2013sinkhorn}. The significance of OT lies in its optimization of a transportation cost matrix that captures cross-modal relationships, with the flexibility to incorporate various modalities and constraints, such as text-text alignment and image-image alignment. Therefore, it aligns the textual and visual features between $Q$ and each retrieved element $R \in \mathcal{R}$ in a fine-grained manner. The Sinkhorn-Knopp algorithm \cite{cuturi2013sinkhorn} further enhances the applicability of OT by introducing entropy regularization, which improves computational efficiency and ensures convergence in large-scale problems. Let $f(U, M)$ represent a scalar that measures the cosine similarity between the question and the retrieved medical report embeddings; $f_{text} = F_{n_q \times n_r}(G^q(t), G^r(t))$ represent a cosine similarity matrix derived from the text embeddings of abnormality descriptions, where $n_q$ is the number of abnormalities in $Q$ and $n_r$ is the number of abnormalities in $R$; and $f_{visual} = F_{n_q \times n_r}(G^q(b), G^r(b))$ represent a cosine similarity matrix derived from the visual features of bounding boxes between $Q$ and $R$. The overall cosine similarity matrix is calculated as follows:
 \begin{equation}
     F(Q, R) = \alpha f(U, M) \cdot \mathbf{1}_{n_q \times n_r} + \beta \cdot f_{text} + \delta \cdot f_{visual},
 \end{equation}
and the transportation cost matrix $C(Q, R)$ is defined as:
\begin{equation}
  C(Q, R) = \mathbf{1}_{n_q \times n_r} - F(Q, R),
\end{equation}
where $\alpha+\beta+\delta=1$ are weights that control the contribution of relevance to the question, similarity in the text and similarity in the visual sense, respectively. This weighting scheme helps adapt to task-specific priorities.

The following Sinkhorn algorithm solves the regularized OT problem:
\begin{equation}
    C_{\gamma}(Q, R) = \min_P \sum_{i, j} P(i, j) \cdot C(Q, R) + \gamma \sum_{i, j} P(i, j) \log P(i, j),
\end{equation}
subject to the constraints: \(\sum_{j} P(i, j) = u_i, \quad \sum_{i} P(i, j) = v_j\)
where $u_i$ and $v_j$ represent the marginal distributions of the query and the retrieved element. The updates for the transportation plan $P$ are iteratively computed as:
\begin{equation}
    P^{(k+1)}(i, j) = u_i \cdot \exp\Big(-\frac{C(Q, R)}{\gamma}\Big) \cdot v_j,
\end{equation}
where $\gamma$ is a regularization parameter controlling the entropy of the distribution. These updates are repeated until convergence to find $P^*$. Hence, the final OT cost between $Q$ and $R$ is computed as: \(\sum_{i, j} P^*(i, j) \cdot C(Q, R)\).

\noindent{\textbf{Answer Generation.}}
After re-ranking, we obtain top-$s$ medical reports $\mathcal{M}=(M_1,\dots, M_s)$ corresponding to the top-ranked elements (i.e., elements with minimal OT cost), where $s<k$. The final input to a frozen VLM consists of $I^q$, $G^q$, $U$ and $\mathcal{M}$ to generate the answer. Algorithm~\ref{alg:method} sums up the proposed methodology.
\begin{algorithm}[t]
\caption{MOTOR}
\label{alg:method}
\textbf{Input:} Query Image $I^q$, Question $U$, Vector Database $\mathcal{D}$ \\
\textbf{Output:} Answer $Ans$ from VLM
\begin{algorithmic}[1]
\State \textbf{Grounded Caption Generation}
    \State \quad Generate $G^q$ for $I^q$
\State \textbf{Retrieval}
    \State \quad Retrieve top-$k$ elements $\mathcal{R} \subset \mathcal{D}$ using cosine similarity between images
\State \textbf{OT-based Re-ranking}
    \State \quad Calculate OT cost for each $R(I^r,G^r,M) \in \mathcal{R}$  w.r.t. $Q(I^q,G^q,U)$
    \State \quad Reorder $\mathcal{R}$ ascending to obtain top $\mathcal{M}$ medical reports
\State \textbf{Answer Generation}
    \State \quad Feed ($Q, \mathcal{M}$) as input to VLM to generate answer $Ans$
    \end{algorithmic}
\end{algorithm}

\section{Experimental Setup}
\noindent{\textbf{Datasets.}}
We utilize MIMIC-CXR-JPG \cite{mimic-cxr,johnson2019mimic-a}, a large-scale external knowledge database that contains chest X-ray images paired with detailed radiology reports. This multimodal dataset serves as a rich resource for tasks such as detecting and localizing abnormalities. The reports were preprocessed by \cite{rameshcxr} to ensure that references to prior reports or images are omitted, which avoids the hallucination of the generative models. For each of the 14 classes in MIMIC-CXR-JPG, we collect 40 images along with their reports from the train split to initialize the vector database. To evaluate our approach, we use the test splits of publicly available MedVQA datasets: Medical-Diff-VQA \cite{hu2023medical} and MIMIC-CXR-VQA \cite{bae2024ehrxqa}. These datasets adhere to the standard VQA format and include diverse image-question-answer pairs. Example questions include: \emph{where is the location of \textit{<abnormality>}?} and \emph{Does the left lung present with any tubes/lines or diseases?}

\noindent{\textbf{Evaluation.}}
Our MedVQA system evaluation protocol employs an automated and human-assisted review. This protocol addresses the challenges posed by open-ended questions, where traditional metrics are not applicable, and the emphasis is placed on factual accuracy \cite{geval}. The automated phase uses an LLM-based approach combined with Natural Language Inference (NLI) to analyze the answer and determine whether the generated answer aligns with or contradicts the ground truth answer. This evaluation approach increases confidence estimation and identifies cases where the model generates correct answers but fails to address all aspects of the question \cite{judgellm,Chiang2023CanLLM,chen-etal-2021-nli-models}. To complement the automated evaluation, a subset of results undergoes manual review by an experienced radiologist to ensure the reliability of the assessment process.

\noindent{\textbf{Implementation Details.}}
We selected different values of $k$ and $s$, and we found that 10 and 5, respectively, achieved the best performance. Visual embeddings are generated using RAD-DINO \cite{perez2024rad} (768D), and text embeddings using ClinicalBERT \cite{alsentzer2019bert} (512D). Re-ranking employs the OT algorithm with $\gamma = 1$ and weights $\alpha = 0.2$, $\beta = 0.3$, and $\delta = 0.5$, identified as the best configuration through extensive experiments. For answer generation, we use open-source state-of-the-art medical VLMs: LLaVA-Med-v1.5-7B \cite{li2024llava} and Dragonfly-Med-v2-8B \cite{chen2024dragonfly}. NLI-based evaluation uses SciFive \cite{phan2021scifive} that is pre-trained on biomedical texts. We use NVIDIA A100 GPU with 40GB memory for all experiments.

\section{Results and Discussion}
\begin{table}[t]
\centering
\caption{Comparison of accuracy(\%) across various retrieval and re-ranking strategies.}
\label{tab:vqa_results}
\setlength{\tabcolsep}{6pt} 
\resizebox{\textwidth}{!}{ 
\begin{tabular}{@{}clcccc@{}}
\toprule
\multirow{2}{*}{\textbf{Model}} & \multirow{2}{*}{\textbf{Method}} & \multirow{2}{*}{\textbf{Reranker Category}} & \multicolumn{2}{c}{\textbf{Dataset}} \\ \cmidrule(lr){4-5}   & & 
& \colorbox{lightorange}{MIMIC-CXR-VQA ↑}
& \colorbox{lightblue}{Medical-Diff-VQA ↑} \\ 
\midrule

\multirow{8}{*}{\rotatebox{90}{\colorbox{promptbg}{\textbf{LLaVA-Med}}}} 
& SG & \multirow{2}{*}{\centering N/A}
& 47.78 & 48.42 \\ 
& MM-RAG, No Re-ranking &
& 51.76 & 47.26 \\ 
\cmidrule(lr){2-5}
& BM25 \cite{rank_bm25} & \multirow{3}{*}{\centering Text-based} &
51.82 & 40.96 \\ 
& LLM-based Reranker \cite{li2023making,team2023gemini} &
& 51.90 & 46.92 \\ 
& BGE M3-Embedding \cite{chen2024bge} & 
& 52.16 & 49.20 \\ 
\cmidrule(lr){2-5}
& Base MM-Reranker & \multirow{2}{*}{\centering Multimodal}
& 50.66 & 57.78 \\ 
& \textbf{MOTOR (ours) } &
& \colorbox{promptbg}{\textbf{56.34}} & \colorbox{promptbg}{\textbf{60.38}} \\ 

\midrule

\multirow{8}{*}{\rotatebox{90}{\colorbox{lightpurple}{\textbf{Dragonfly-Med}}}} 
& SG & \multirow{2}{*}{\centering N/A}
& 56.00 & 47.10 \\ 
& MM-RAG, No Re-ranking &
& 56.40 & 52.84 \\ 
\cmidrule(lr){2-5}
& BM25 \cite{rank_bm25} & \multirow{3}{*}{\centering Text-based}
& 56.40 & 63.11 \\ 
& LLM-based Reranker \cite{li2023making,team2023gemini}&
& 55.66 & 63.46 \\ 
& BGE M3-Embedding \cite{chen2024bge} &
& 55.68 & 61.24 \\ 
\cmidrule(lr){2-5}
& Base MM-Reranker & \multirow{2}{*}{\centering Multimodal}
& 54.78 & 61.84 \\ 
& \textbf{MOTOR (ours) } &
& \colorbox{lightpurple}{\textbf{58.04}} & \colorbox{lightpurple}{\textbf{64.52}} \\ 

\bottomrule
\end{tabular}
} 
\end{table}

\begin{figure}[htbp!]
    \centering
    \includegraphics[width=\textwidth]{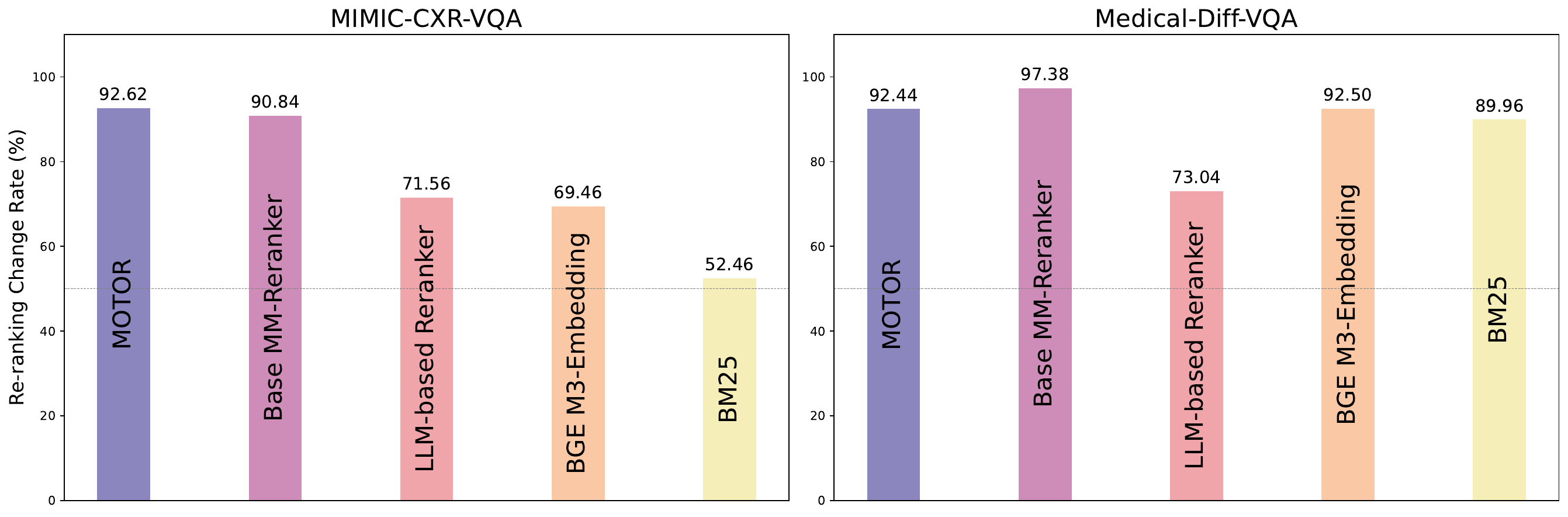}
    \caption{Proportion of samples with modified retrieved context after re-ranking.}
    \label{fig:rate}
\end{figure}


The empirical results in Table~\ref{tab:vqa_results} demonstrate the effectiveness of MOTOR in enhancing the factual accuracy between generated and ground truth answers by re-ordering retrievals to prioritize more relevant ones that were initially underrated. 

First, the comparison between SG and MM-RAG reveals a key trade-off. SG underperforms when the external context is needed, while MM-RAG struggles with noisy retrieved context, degrading performance. This underscores the challenge of balancing external knowledge inclusion with retrieval precision.

Second, text-based re-ranking methods such as BM25 \cite{rank_bm25}, BGE M3-Embedding \cite{chen2024bge} and the LLM-based reranker (Gemma 2B) \cite{li2023making,team2023gemini} perform better than MM-RAG without re-ranking on the MIMIC-CXR-VQA dataset but underperform on Medical-Diff-VQA, despite the higher re-ranking change rates in the latter (i.e., the number of samples having their initial retrievals modified divided by the total number of samples), as shown in Fig.~\ref{fig:rate}. This inconsistency reflects the limitations of sparse lexical matching in complex multimodal tasks and the bias toward textual similarity over visual-textual relationships. The results emphasize the need for multimodal re-ranking strategies that integrate both modalities effectively.

\begin{figure}[htbp!]
    \centering
    \includegraphics[width=0.7\textwidth]{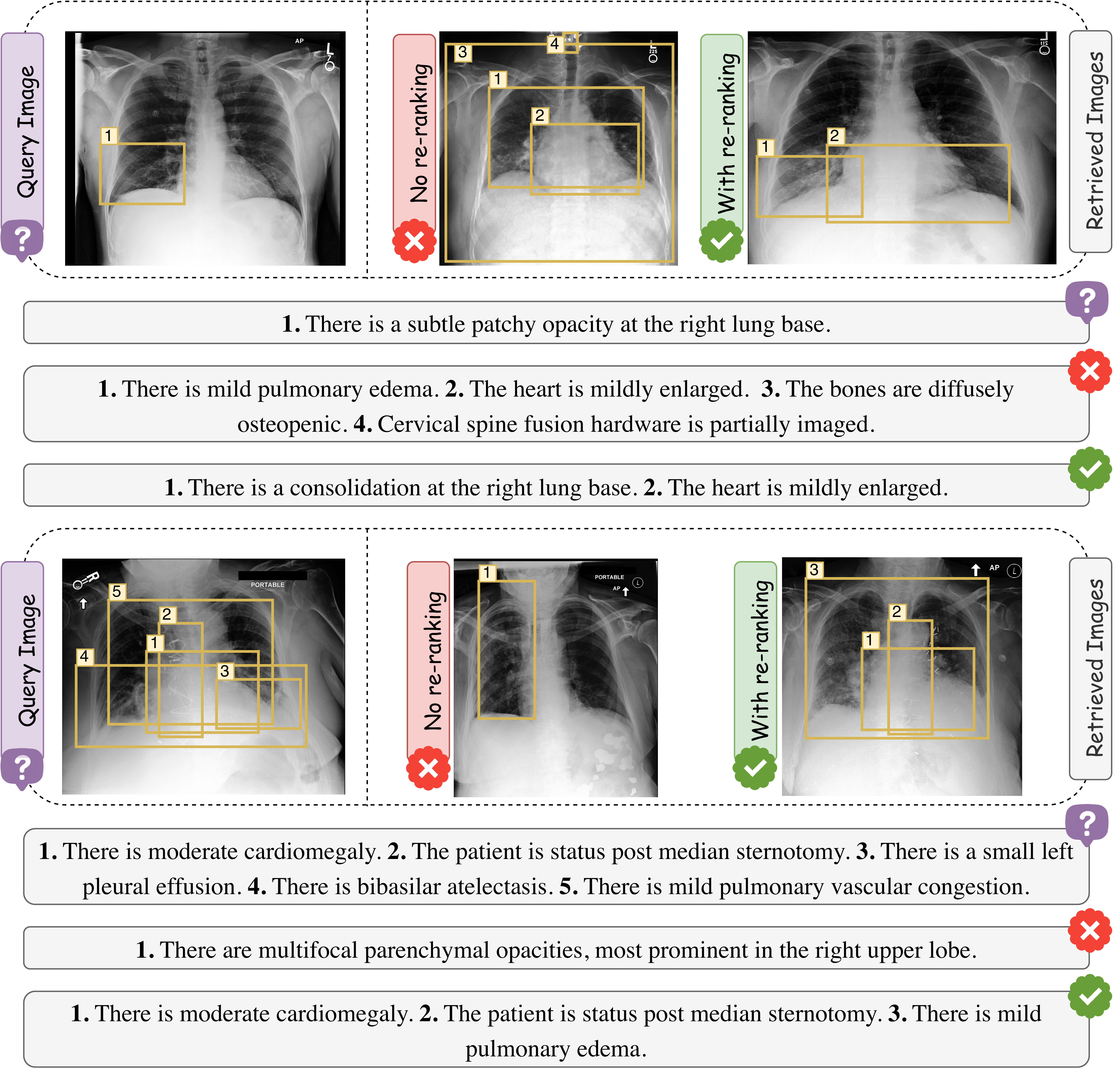}
    \caption{Comparison of the clinical relevance of retrievals: before and after re-ranking.}
    \label{fig:q-results}
\end{figure}

Additionally, MOTOR achieves state-of-the-art accuracy in both datasets, outperforming SG and all other MM-RAG strategies, including the base multimodal reranker (Base MM-Reranker) with an average improvement of 6.45\% (+3.77\% on MIMIC-CXR-VQA and +9.12\% on Medical-Diff-VQA). This improvement validates MOTOR's ability to filter irrelevant noise while prioritizing contextually relevant multimodal information. The contrast between MOTOR and Base MM-Reranker, which uses cosine similarity without OT, highlights the limitations of shallow multimodal integration. MOTOR's success indicates the necessity of fine-grained, structured optimization techniques for robust retrieval in MedVQA tasks without extra memory. The extra cost at inference time is limited to cosine similarity computations and Sinkhorn-based OT (worst case: $O(n_q \cdot n_r)$) \cite{cuturi2013sinkhorn}. Moreover, qualitative results from Fig.~\ref{fig:q-results} show how MOTOR is capable of retrieving more precise medical context, which was underrated in the initial retrieval. In addition, we conducted an ablation study on different OT weighting configurations, either fully prioritizing one modality or balancing visual and textual contributions. Results indicate that accuracy remains comparable between approaches that emphasize textual similarity (while still incorporating visual similarity) and those that emphasize visual similarity (while still considering textual similarity). The average accuracy was 54.54\% for text-prioritized retrieval and 54.48\% for visual-prioritized retrieval. The minimum accuracy observed was 53.08\% for the text-prioritized approach and 53.48\% for the visual-prioritized approach, while the maximum reached 56.02\% and 56.34\%, respectively. These findings suggest that both modalities contribute equally to performance, with no significant advantage of prioritizing one over the other. We also conducted ablations on different values of $k$ and $s$ and found that performance changes were minimal (±0.2\%).

Furthermore, to assess the quality of the generated grounded captions, an expert radiologist independently reviewed randomly selected images along with their corresponding generated captions. The findings indicate that around 74\% of these captions were verified to be accurate, demonstrating strong alignment with expert interpretations. In addition, we asked a radiologist to blindly review a random subset of query images alongside their top retrieved candidates---both in original and re-ranked order. As shown in Fig.~\ref{fig:q-results}, re-ranking effectively promoted images with higher abnormality matches to top positions while demoting those with mismatches. This confirms that the re-ranking prioritizes clinically relevant cases.

Finally, to verify the robustness of the evaluation method, we performed a manual review of a subset of samples and found that over 98\% of the cases yielded a meaningful evaluation prediction. This confirms the reliability of the metric used to evaluate open-ended MedVQA systems.

\section{Conclusion}
This study proposed MOTOR, a novel multimodal approach that establishes a new benchmark for retrieval-augmented MedVQA systems, achieving performance gains through multimodal feature alignment. This reduces noise in retrieved contexts, and enriches the VLM’s access to task-relevant information. MOTOR addressed the limitations of text-based re-ranking strategies by modelling an advanced mechanism for multimodal re-ranking. The alignment of clinical relevance between the query and the retrieved context through grounded captions and OT empowered the retrieval quality, thereby enhancing the factual correctness in MedVQA. Future work could explore adapting MOTOR to other medical imaging modalities and tasks, such as medical report generation.

\begin{credits}
\subsubsection{\discintname}
The authors have no competing interests to declare that are relevant to the content of this article.
\end{credits}

%
%
%
\bibliographystyle{splncs04}
\bibliography{references}

\begin{thebibliography}{10}
\providecommand{\url}[1]{\texttt{#1}}
\providecommand{\urlprefix}{URL }
\providecommand{\doi}[1]{https://doi.org/#1}

\bibitem{alsentzer2019bert}
Alsentzer, E., et~al.: Publicly available clinical {BERT} embeddings. In: Rumshisky, A., Roberts, K., Bethard, S., Naumann, T. (eds.) Proceedings of the 2nd Clinical Natural Language Processing Workshop. pp. 72--78. Association for Computational Linguistics, Minneapolis, Minnesota, USA (Jun 2019). \doi{10.18653/v1/W19-1909}

\bibitem{bae2024ehrxqa}
Bae, S., et~al.: Ehrxqa: A multi-modal question answering dataset for electronic health records with chest x-ray images. Advances in Neural Information Processing Systems  \textbf{36} (2024)

\bibitem{bannur2024maira}
Bannur, S., et~al.: Maira-2: Grounded radiology report generation. arXiv preprint arXiv:2406.04449  (2024)

\bibitem{rank_bm25}
Brown, D.: {Rank-BM25: A Collection of BM25 Algorithms in Python} (2020). \doi{10.5281/zenodo.4520057}

\bibitem{chen2024bge}
Chen, J., et~al.: {M}3-embedding: Multi-linguality, multi-functionality, multi-granularity text embeddings through self-knowledge distillation. In: Ku, L.W., Martins, A., Srikumar, V. (eds.) Findings of the Association for Computational Linguistics: ACL 2024. pp. 2318--2335. Association for Computational Linguistics, Bangkok, Thailand (Aug 2024). \doi{10.18653/v1/2024.findings-acl.137}

\bibitem{chen-etal-2021-nli-models}
Chen, J., et~al.: Can {NLI} models verify {QA} systems' predictions? In: Moens, M.F., Huang, X., Specia, L., Yih, S.W.t. (eds.) Findings of the Association for Computational Linguistics: EMNLP 2021. pp. 3841--3854. Association for Computational Linguistics, Punta Cana, Dominican Republic (Nov 2021). \doi{10.18653/v1/2021.findings-emnlp.324}

\bibitem{chen2024dragonfly}
Chen, K., et~al.: Dragonfly: Multi-resolution zoom supercharges large visual-language model. arXiv e-prints pp. arXiv--2406 (2024)

\bibitem{chen2022multi}
Chen, Z., et~al.: Multi-modal masked autoencoders for medical vision-and-language pre-training. In: International Conference on Medical Image Computing and Computer-Assisted Intervention. pp. 679--689. Springer (2022)

\bibitem{Chiang2023CanLLM}
Chiang, C.H., et~al.: Can large language models be an alternative to human evaluations? In: Annual Meeting of the Association for Computational Linguistics (2023)

\bibitem{cuturi2013sinkhorn}
Cuturi, M.: Sinkhorn distances: Lightspeed computation of optimal transport. Advances in neural information processing systems  \textbf{26} (2013)

\bibitem{ragsurvey}
Fan, W., et~al.: A survey on rag meeting llms: Towards retrieval-augmented large language models. In: Proceedings of the 30th ACM SIGKDD Conference on Knowledge Discovery and Data Mining. p. 6491–6501. KDD '24, Association for Computing Machinery, New York, NY, USA (2024). \doi{10.1145/3637528.3671470}

\bibitem{gao2023retrieval}
Gao, Y., et~al.: Retrieval-augmented generation for large language models: A survey. arXiv preprint arXiv:2312.10997  (2023)

\bibitem{ragreview}
Hartsock, I., et~al.: Vision-language models for medical report generation and visual question answering: a review. Frontiers in Artificial Intelligence  \textbf{7} (2024). \doi{10.3389/frai.2024.1430984}

\bibitem{hu2023medical}
Hu, X., et~al.: Medical-diff-vqa: a large-scale medical dataset for difference visual question answering on chest x-ray images. PhysioNet  \textbf{12}, ~13 (2023)

\bibitem{mimic-cxr}
Johnson, A.E.W., et~al.: Mimic-cxr, a de-identified publicly available database of chest radiographs with free-text reports. Scientific data.  \textbf{6}(1) (2019-12-12)

\bibitem{johnson2019mimic-a}
Johnson, A.E., et~al.: Mimic-cxr-jpg, a large publicly available database of labeled chest radiographs. arXiv preprint arXiv:1901.07042  (2019)

\bibitem{lewis2020retrieval}
Lewis, P., et~al.: Retrieval-augmented generation for knowledge-intensive nlp tasks. Advances in Neural Information Processing Systems  \textbf{33},  9459--9474 (2020)

\bibitem{li2023making}
Li, C., et~al.: Making large language models a better foundation for dense retrieval. arXiv preprint arXiv:2312.15503  (2023)

\bibitem{li2024llava}
Li, C., et~al.: Llava-med: Training a large language-and-vision assistant for biomedicine in one day. Advances in Neural Information Processing Systems  \textbf{36} (2024)

\bibitem{lin2023medical}
Lin, Z., et~al.: Medical visual question answering: A survey. Artificial Intelligence in Medicine  \textbf{143},  102611 (2023)

\bibitem{geval}
Liu, Y., et~al.: {G}-eval: {NLG} evaluation using gpt-4 with better human alignment. In: Bouamor, H., Pino, J., Bali, K. (eds.) Proceedings of the 2023 Conference on Empirical Methods in Natural Language Processing. pp. 2511--2522. Association for Computational Linguistics, Singapore (Dec 2023). \doi{10.18653/v1/2023.emnlp-main.153}

\bibitem{perez2024rad}
P{\'e}rez-Garc{\'\i}a, F., et~al.: {Exploring scalable medical image encoders beyond text supervision}. Nature Machine Intelligence  \textbf{7}(1),  119--130 (2025). \doi{10.1038/s42256-024-00965-w}

\bibitem{phan2021scifive}
Phan, L.N., et~al.: Scifive: a text-to-text transformer model for biomedical literature. arXiv preprint arXiv:2106.03598  (2021)

\bibitem{rameshcxr}
Ramesh, V., et~al.: Cxr-pro: Mimic-cxr with prior references omitted  (2022)

\bibitem{shaaban2024medpromptx}
Shaaban, M.A., et~al.: Medpromptx: Grounded multimodal prompting for chest x-ray diagnosis. arXiv preprint arXiv:2403.15585  (2024)

\bibitem{shi2024survey}
Shi, C., et~al.: A survey on trustworthiness in foundation models for medical image analysis. arXiv preprint arXiv:2407.15851  (2024)

\bibitem{team2023gemini}
Team, G., et~al.: Gemini: a family of highly capable multimodal models. arXiv preprint arXiv:2312.11805  (2023)

\bibitem{thirunavukarasu2023large}
Thirunavukarasu, A.J., et~al.: Large language models in medicine. Nature medicine  \textbf{29}(8),  1930--1940 (2023)

\bibitem{villani2009optimal}
Villani, C., et~al.: Optimal transport: old and new, vol.~338. Springer (2009)

\bibitem{wang2022medclip}
Wang, Z., et~al.: {M}ed{CLIP}: Contrastive learning from unpaired medical images and text. In: Goldberg, Y., Kozareva, Z., Zhang, Y. (eds.) Proceedings of the 2022 Conference on Empirical Methods in Natural Language Processing. pp. 3876--3887. Association for Computational Linguistics, Abu Dhabi, United Arab Emirates (Dec 2022). \doi{10.18653/v1/2022.emnlp-main.256}

\bibitem{xia2024rule}
Xia, P., et~al.: Rule: Reliable multimodal rag for factuality in medical vision language models. In: Proceedings of the 2024 Conference on Empirical Methods in Natural Language Processing. pp. 1081--1093 (2024)

\bibitem{yang2023impact}
Yang, J., et~al.: The impact of chatgpt and llms on medical imaging stakeholders: perspectives and use cases. Meta-Radiology p. 100007 (2023)

\bibitem{ramm}
Yuan, Z., et~al.: Ramm: Retrieval-augmented biomedical visual question answering with multi-modal pre-training. p. 547–556. MM '23, Association for Computing Machinery, New York, NY, USA (2023). \doi{10.1145/3581783.3611830}

\bibitem{zhang2023pmc}
Zhang, X., et~al.: Pmc-vqa: Visual instruction tuning for medical visual question answering. arXiv preprint arXiv:2305.10415  (2023)

\bibitem{judgellm}
Zheng, L., et~al.: Judging llm-as-a-judge with mt-bench and chatbot arena. In: Proceedings of the 37th International Conference on Neural Information Processing Systems. NIPS '23, Curran Associates Inc., Red Hook, NY, USA (2023)

\end{thebibliography}

\end{document}